\documentclass[preprint,12pt]{elsarticle}

\usepackage{amssymb}
\usepackage{amsmath}
\usepackage{lineno}
\usepackage{caption}
\usepackage{here}
\journal{Neural Networks}

\title{Investigating Fine- and Coarse-grained Structural Correspondences Between Deep Neural Networks and Human Object Image Similarity Judgments Using Unsupervised Alignment}

\author[1]{Soh Takahashi\fnref{fn1}}
\author[1]{Masaru Sasaki\fnref{fn1}}
\author[1]{Ken Takeda}
\author[1]{Masafumi Oizumi\corref{cor1}} 

\affiliation[1]{organization={Graduate School of Arts and Science, University of Tokyo},%Department and Organization
            addressline={3-8-1 Komaba}, 
            city={Meguro-ku},
            postcode={153-8902}, 
            state={Tokyo},
            country={Japan}}

\fntext[fn1]{These authors contributed equally to this work}
\cortext[cor1]{Corresponding author: c-oizumi@g.ecc.u-tokyo.ac.jp}

\begin{document}

\begin{frontmatter}

\begin{abstract}
The learning mechanisms by which humans acquire internal representations of objects are not fully understood. Deep neural networks (DNNs) have emerged as a useful tool for investigating this question, as they have internal representations similar to those of humans as a byproduct of optimizing their objective functions. While previous studies have shown that models trained with various learning paradigms—such as supervised, self-supervised, and CLIP—acquire human-like representations, it remains unclear whether their similarity to human representations is primarily at a coarse category level or extends to finer details. Here, we employ an unsupervised alignment method based on Gromov-Wasserstein Optimal Transport to compare human and model object representations at both fine-grained and coarse-grained levels. The unique feature of this method compared to conventional representational similarity analysis is that it estimates optimal fine-grained mappings between the representation of each object in human and model representations. We used this unsupervised alignment method to assess the extent to which the representation of each object in humans is correctly mapped to the corresponding representation of the same object in models. Using human similarity judgments of 1,854 objects from the THINGS dataset, we find that models trained with CLIP consistently achieve strong fine- and coarse-grained matching with human object representations. In contrast, self-supervised models showed limited matching at both fine- and coarse-grained levels, but still formed object clusters that reflected human coarse category structure. Our results offer new insights into the role of linguistic information in acquiring precise object representations and the potential of self-supervised learning to capture coarse categorical structures.
\end{abstract}

% \begin{graphicalabstract}
%\includegraphics{grabs}
% \end{graphicalabstract}

\begin{highlights}
\item Introduces an unsupervised alignment to assess human-like object representations of DNNs.
\item CLIP models show highest fine-grained matching (20\% top-1 match).
\item This underscores the role of linguistic cues in refining representations.
\item Image-only self-supervised models lack fine matching with human representations.
\item Rather, they capture coarse category structures, hinting at prelinguistic links.
\end{highlights}

\begin{keyword}
Deep Neural Networks \sep 
Human Object Representations \sep
Unsupervised Alignment \sep
Gromov-Wasserstein Optimal Transport \sep
Representational Similarity Analysis
\end{keyword}

\end{frontmatter}

% \linenumbers
\section{Introduction}
To understand how humans develop internal visual representations of objects, recent studies have compared human representations with those of deep neural networks (DNNs). Broadly speaking, these comparisons have taken two distinct approaches, neural response-based methods and behavioral response-based methods. The neural response-based methods investigate correspondences between DNN activations and neural responses measured through techniques such as fMRI and electrophysiology \cite{Yamins2013-av, Yamins2014-qn, Khaligh-Razavi2014-yr, Cadieu2014-bm, Cichy2016-ga, Kietzmann2019-hn, Bakhtiari2021-cb, Zhuang2021-ek, Xu2021-cz, Konkle2022-tx, Prince2024-zf}. In contrast, the behavioral response-based methods compare DNN activations with human psychological representations derived from psychological experiments \cite{Kheradpisheh2016-wd, Groen2018-mz, Peterson2018-ie, Bracci2019-qy, King2019-pv, muttenthaler2023-zf}.

Recent advance in psychological research have enabled the collection of large-scale datasets containing numerous objects \cite{Hebart2020-gr, Hebart2023-im, Roads2021-vq}, making it feasible to conduct detailed structural comparisons between human psychological representations and DNN representations. Here, using these extensive psychological datasets, this study systematically examined structural alignments between DNN and human psychological representation.

Against this background, one promising approach to exploring potential learning mechanisms in the human brain is to compare DNNs trained with different learning paradigms to human representations and assess their similarities and differences. To this end, recent studies have focused on three major learning paradigms: supervised learning on labeled image data, self-supervised learning from unlabeled image data, and vision-language learning integrating visual and linguistic modalities. Supervised learning mirrors human learning when object names are explicitly taught. However, human development involves learning beyond labeled data, such as recognizing perceptual patterns without labels, which may be critical in early childhood before language acquisition \cite{Saffran2018-lr, Cusack2024-yw}. Image-based self-supervised learning with unlabeled data offers a potential computational framework to capture this aspect and is gaining attention \cite{Lotter2020-iv,Zhuang2021-ek,Konkle2022-tx,kataoka2025-ei}. Another key aspect of human learning is associating visual and linguistic information without explicit supervision \cite{Bergelson2012-lu,Frank2017-nb}. Vision-language models like CLIP \cite{Radford2021-ky}, which learn correspondences between visual and textual data, may capture this process. Given these connections to human learning processes, we focused on these three learning paradigms in this study.

Although findings vary, studies agree that no single learning paradigm produces representations significantly closer to those of humans; instead, each paradigm exhibits some degree of similarity. Pioneering studies have demonstrated that supervised learning models, trained on classification tasks, resemble representations in the inferior temporal (IT) cortex \cite{Yamins2013-av, Yamins2014-qn, Khaligh-Razavi2014-yr, Storrs2021-ze} and reflect hierarchical structures observed in human visual processing \cite{Cadieu2014-bm}. Compared to supervised learning, image-based self-supervised learning has been considered more biologically plausible, as it does not rely on large-scale labeled data. These models predict neural activity in the visual pathways as effectively as, and in some cases even better than, supervised learning models \cite{Zhuang2021-ek, Konkle2022-tx, Bakhtiari2021-cb} and naturally develop category-selective features that correspond to human brain activity \cite{Prince2024-zf}. They have also been shown to capture aspects of human behavior, such as visual illusions \cite{Lotter2020-iv} and gloss perception \cite{Storrs2021-al}. Unlike these unimodal approaches, CLIP has been suggested to be a potential model for vision-language association learning in humans. For example, using only the correspondences between visual frames and child-directed utterances in head-mounted camera recordings of a single child aged 6 to 25 months, CLIP acquired word-referent mappings and generalized to novel visual objects \cite{Vong2024-gx}. Moreover, CLIP-trained models show stronger correlations with human fMRI responses and psychological representations \cite{Muttenthaler2021-bp}, better predict fMRI data \cite{Wang2023-de}, and more closely mimic human behavior in object recognition tasks, including both performance and error patterns \cite{Geirhos2021-io, muttenthaler2023-zf, Demircan2023-gs}. In addition, a recent large-scale study using Representational Similarity Analysis (RSA) demonstrated that DNNs trained with these diverse learning objectives exhibit comparable similarity to human visual representations \cite{Conwell2024-mp}.

\begin{figure}[tp!]
\centering
\includegraphics[width=\textwidth]{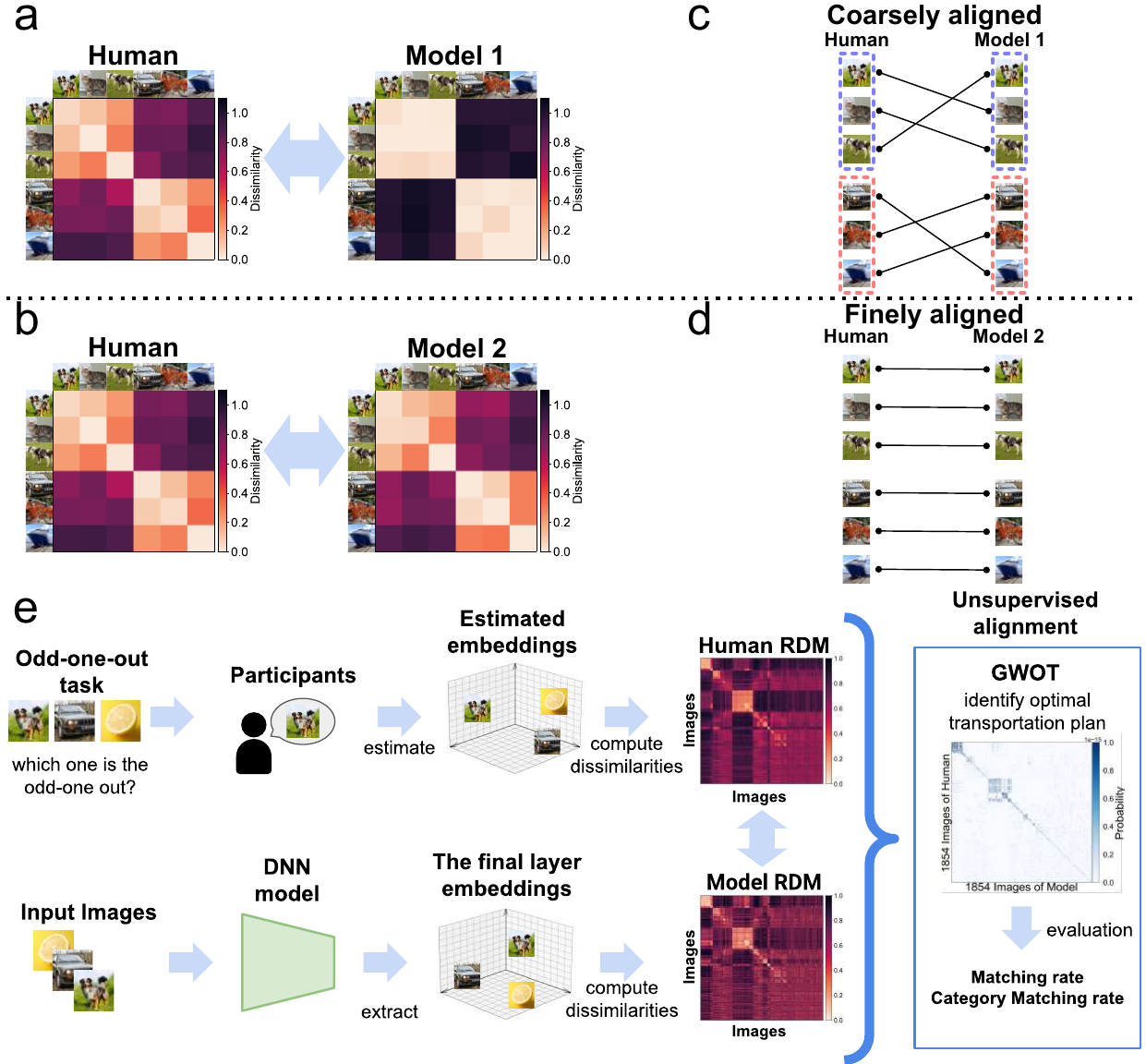}
\caption{\textbf{Overview of alignment methods and experimental workflow.}
\textbf{a.b.} Schematic of human and model object representation dissimilarities (RDMs). 
\textbf{c.d.} Schematic of object representation mapping between human and model.
\textbf{e.} Workflow: Human RDMs were derived from behavioral odd-one-out tasks, while model RDMs were computed from DNN final layer embeddings. Unsupervised alignment was performed using GWOT and evaluated by the matching rate metric.}
\label{fig:schematic}
\end{figure}

However, the question of whether the similarity between DNNs trained with each learning paradigm and human representations exists primarily at a coarse-grained level (e.g., distinguishing broad object categories like animals and vehicles only) or a fine-grained level (e.g., differentiating specific types of animals or vehicles, akin to human discrimination) remains largely unexplored. As one example of this issue, Figures \ref{fig:schematic}a and b illustrate a schematic comparison of humans and model representational dissimilarity matrices (RDMs). In this example, Model 1 primarily captures dissimilarities between broad categories (e.g., animals vs. vehicles) but fails to preserve finer distinctions within categories which typically humans have. In contrast, Model 2 captures more fine-grained dissimilarities within categories (e.g., among animal species or vehicle types) and more closely resembles human representations than Model 1. However, conventional RSA, widely used to compare human and model representations, yields similarly high correlations for both Model 1 and Model 2 with human representations, and thus fails to distinguish their qualitative differences. This limitation stems from RSA's implicit assumption of a "fixed mapping" between human and model objects—for instance, mapping a human’s representation of a dog to a model’s representation of a dog when computing correlations (see \cite{takeda2025-qk, Takeda2025-mc} for more detailed explanations). We term this approach “supervised” comparison (distinct from the “supervised” learning paradigm in DNN training). To evaluate whether models exhibit representational structures aligned with human representations at the fine-grained level, it is necessary to move beyond the fixed mapping assumption of supervised comparison.

To investigate whether the similarity between DNNs and humans lies at the coarse or fine-grained level, we employed an unsupervised alignment method that estimates the mapping itself between human and model representations. Unlike supervised comparison, unsupervised alignment estimates the mapping between human and model objects based on their internal distance structures. Figures \ref{fig:schematic}c and d illustrate the estimated mappings in the schematic example. For Model 1, the mapping preserves category-level correspondence, but not object-level correspondence. That is, the mapping is coarse-grained but not fine-grained. In contrast, the mapping between Model 2 and human representations preserves object-level correspondence, indicating fine-grained matching. This example shows that unsupervised alignment can distinguish model differences that supervised comparison overlooks. To conduct unsupervised alignment, we adopt Gromov-Wasserstein Optimal Transport (GWOT) \cite{Memoli2011-rf, Peyre2016-ek}, which has been successfully applied to various contexts \cite{Alvarez-Melis2018-vf, Demetci2021-xv, kawakita2023-wa, Takeda2025-mc, takeda2025-qk, Kawakita2025-mv}. 

Using this unsupervised alignment framework, we examined how models trained with different learning paradigms align with human psychological representations at fine-grained and coarse-grained levels. We utilized the THINGS dataset \cite{Hebart2019-th, Hebart2020-gr, Hebart2023-im}, which includes human similarity judgments for 1,854 naturalistic object images grouped into broader categories. To assess the alignment between human and model object representations, we applied an unsupervised alignment framework based on GWOT. At the fine-grained level, our results showed that CLIP-based models consistently demonstrated the strongest matching with human representations, while self-supervised models performed at chance levels. At the coarse-grained level, linguistic models, including CLIP and supervised variants, showed robust matching, while self-supervised models, though less effective than linguistic models, exceeded chance levels. Clustering analysis revealed that the coarse, human-like clusters in self-supervised models account for their modest coarse-grained alignment, comparable to clusters formed by linguistic models. These findings highlight the critical role of linguistic information in achieving precise representational alignment and emphasize the potential of self-supervised learning to capture coarse categorical structures.

\section{Methods}
\subsection{Obtaining Object Representational Structures from humans and DNNs}
To capture human object representational structures, we utilized the THINGS dataset, which contains 4.7 million similarity judgments for 1,854 naturalistic objects \cite{Hebart2019-th,Hebart2020-gr,Hebart2023-im}. These similarity judgments were obtained through an odd-one-out task, where participants selected the image that was most dissimilar to the others from a set of three object images. The naturalistic objects in the dataset are classified into 27 coarse categories by human raters in the dataset. After removing subcategories of other categories and categories with fewer than 10 concepts, we used 20 coarse categories in this study: animal, plant, body part, clothing, clothing accessory, food, drink, container, furniture, home decor, tool, office supply, electronic device, medical equipment, musical instrument, vehicle, part of car, sports equipment, toy, and weapon. These categories include 1,283 concepts. 

To model similarity judgments derived from odd-one-out tasks, we estimated low-dimensional embeddings for 1,854 objects using SPoSE (Similarity-based Projective Sparse Embedding), as previously proposed \cite{Zheng2019-dv, Hebart2020-gr}. SPoSE optimizes non-negative embeddings to capture human-perceived similarity through a cross-entropy loss. In our implementation, embeddings were initialized with 70 latent dimensions, and similarities were computed using the dot product, as reported \cite{Hebart2020-gr}. We applied a penalty term to encourage the L2 norm of each embedding vector to be close to 1, enhancing training stability. The regularization parameter was selected via Optuna \cite{Akiba2019-jc} within the range [$10^{-4}, 10^{^1}$] (log scale) over 50 trials. The model was optimized using the Adam algorithm with a learning rate of 0.001, a minibatch size of 2,048 triplets, and 40 epochs. Using the estimated embeddings, we computed RDMs based on dot products.

To capture the object representational structures of Deep Neural Network (DNN) models, we used the same 1,854 object images from the THINGS dataset as inputs. We extracted embeddings from the each model's final latent layers, capturing their learned object representations. Using these embeddings, we computed RDMs based on cosine distances. Figure \ref{fig:schematic}b illustrates the process of obtaining object representational structures from humans and DNNs and aligning them.

\subsection{Conventional Representational Similarity Analysis (RSA)}
To compare object representational structures between DNNs and humans using supervised comparison, we calculated correlations between RDMs derived from human and model embeddings. This approach, known as conventional Representational Similarity Analysis (RSA) \cite{Kriegeskorte2008-rv}, is widely used to quantify similarity between representational structures across systems. Specifically, we computed the Pearson correlation coefficients by flattening the upper triangular parts of the RDMs into one-dimensional arrays and computing the correlation between them. It is important to note that this method assumes a fixed mapping between objects, i.e., mapping between the same objects, in the two systems.

\subsection{Unsupervised alignment framework based on Gromov-Wasserstein Optimal Transport}
To compare structural relationships in object representations between DNNs and humans using an unsupervised approach, we applied Gromov-Wasserstein Optimal Transport (GWOT). GWOT is an unsupervised alignment method which finds the optimal transportation plan between two metric spaces by aligning their intrinsic structural relationships \cite{Memoli2011-rf}. The objective function of GWOT is defined as:

\[
{\rm GWD}=\min_{\Gamma} \left[\sum_{i,j,k,l} (D_{ij}-D_{kl}')^2\Gamma_{ik}\Gamma_{jl} \right]
\]

where $D$ and $D'$ represent the pairwise dissimilarities within the source and target domains, respectively, and $\Gamma$ is a transportation between the two domains. The transportation plan $\Gamma$ is optimized under the following constraints: $\sum_j \Gamma_{ij}=p_i$, $\sum_i \Gamma_{ij}=q_j$, and $\sum_{ij} \Gamma_{ij}=1$, where $p$ and $q$ are the distributions of mass in the source and target domains. For simplicity, we set $p$ and $q$ as uniform distributions. Each entry $\Gamma_{ij}$ represents the mass transported from the $i$-th point in the source domain to the $j$-th point in the target domain.

\subsubsection{Hyperparameter tuning}
To efficiently identify good local optima, we employed an entropy-regularized version of GWOT, known as entropic GWOT \cite{Peyre2016-ek, Peyre2019-bl}. The entropic GWOT objective function is defined as:
\[
{\rm GWD}_{\varepsilon} = \min_{\Gamma} \left[ \sum_{i,j,k,l} (D_{ij}-D_{kl}')^2 \Gamma_{ik}\Gamma_{jl} - \varepsilon H(\Gamma) \right]
\]
where $H(\Gamma)$ is the entropy of the transportation plan $\Gamma$, and $\varepsilon$ is a hyperparameter which controls the trade-off between the original GWOT objective and the entropy term. The original GWOT formulation is a special case when $\varepsilon = 0$. By appropriately tuning $\varepsilon$, the entropic GWOT ensures smoother transportation plans and better exploration of the solution space, enhancing both efficiency and accuracy. 

To find good local optimums, we conducted hyperparameter tuning on $\varepsilon$ by using the GWTune toolbox we previously developed \cite{Takeda2025-mc}. This toolbox uses Optuna \cite{Akiba2019-jc} for hyperparameter tuning and Python Optimal Transport (POT) \cite{Flamary2021-hd} for GWOT optimization. We sampled 500 different values of $\varepsilon$ ranging from 0.0001 to 0.001 by a Bayesian sampler called TPE (Tree-structured Parzen Estimator) sampler \cite{Bergstra2013-wt}. We chose the value of $\varepsilon$, where the optimal transportation plan minimizes the Gromov-Wasserstein distance without the entropy-regularization term following a procedure proposed in a previous study \cite{Demetci2022-ba}.

\subsubsection{Initialization of transportation plan}
To avoid poor local minima, we randomly initialized the transportation plan multiple times, as previously suggested \cite{Takeda2025-mc}. Each element in the initial transportation plan was sampled from the uniform distribution [0,1] and was normalized to satisfy the following conditions: 
$\sum_j \Gamma_{ij}=p_i$, $\sum_i \Gamma_{ij}=q_j$, and $\sum_{ij} \Gamma_{ij}=1$. For each value of $\varepsilon$, the plan was randomly initialized.

\subsubsection{Metrics for evaluating unsupervised alignment}
To evaluate how well the object representational structures align between human and DNN embeddings at the fine-grained level, we calculated a metric called the "matching rate". For each object $i$ in the source domain, we determined whether the highest mass in the corresponding row of the transportation plan $\Gamma$ aligns with its correct counterpart in the target domain:
\[
\text{{Match}}(i) = \begin{cases} 
1, & \text{if } \underset{j}{\operatorname{argmax}} (\Gamma_{ij})=i\\
0, & \text{otherwise}.
\end{cases}
\]
The matching rate is the percentage of correctly matched objects:
\[
\text{{Matching rate}} = \frac{{\sum_{i=1}^{n} \text{{Match}}(i)}}{n}.
\]

To evaluate coarse-grained alignment, we also computed the “category matching rate.”. For each object $i$ in the source domain, we assessed whether the highest mass in the corresponding row of the transportation plan $\Gamma$ aligns with the correct category in the target domain, rather than an individual object. Specifically, we defined a match as follows:
\[
\text{{Category Match}}(i) = \begin{cases} 
1, & \text{if } \underset{j}{\operatorname{argmax}} (\Gamma_{ij}) \in C \\
0, & \text{otherwise}.
\end{cases}
\]
The category matching rate is then calculated as the proportion of objects for which the correct category is identified:
\[
\text{{Category matching rate}} = \frac{{\sum_{i=1}^{n} \text{{Category Match}}(i)}}{n}.
\]
This metric allows us to assess the degree to which the overall structure of object categories in the DNN embeddings corresponds to the human categorization structure.

We calculate these metrics for all transportation plans obtained in the optimization process and report the results for the transportation plan with the lowest GWD and the transportation plan with the highest (category) matching rate. The reason for reporting the latter (highest matching rate) transportation plan is that the matching rate is a highly discontinuous metric. In a finite optimization process (500 iterations in this study), the solution minimizing GWD may not consistently reflect the overall trend in matching rate and may be susceptible to random variations. Thus, reporting the plan with the highest matching rate complements the GWD-minimized solution by providing additional insight into the metrics' behaviors.

\subsubsection{Estimating chance-level matching rates through simulation}
To evaluate the chance-level of the matching rate and category matching rate, we generated random transportation plans and computed the corresponding matching rates. For simplicity, the transportation plans were generated as binary matrices, where each element was either 0 or 1.

For a basic estimation of chance level, we randomly generated 1,000 transportation plans and computed the mean, 5th percentile, and 95th percentile of the resulting matching rates and category matching rates.

In our analysis, we also report the highest matching rate within the iterative optimization process of the unsupervised alignment. To assess its chance level, we simulated this selection procedure as follows. We generated a transportation plan for each of the 500 iterations performed in the unsupervised alignment process, then recorded the highest matching rate and highest category-level matching rate obtained in each simulation. This procedure was repeated 1,000 times, and we computed the mean, 5th percentile, and 95th percentile of the resulting values.

\subsection{DNN learning objectives}
To investigate how different learning paradigms capture human-like object representational structures, we analyzed a variety of publicly available DNN models. Table \ref{tab:models} lists the models, including their learning paradigms, architectures, and training datasets. Due to the computational cost of GWOT optimization for evaluating representational similarity, we examined 5 to 8 models per learning paradigm. The learning paradigms considered were as follows:

\subsubsection{Supervised learning}
Supervised learning trains a model to predict object labels provided by a teacher. The objective function is:
\[
\mathcal{L}(\theta) = - \sum_{i=1}^{N} \sum_{c=1}^{C} y_{i,c} \log \hat{y}_{i,c}
\]
where \( y_{i,c} \) is the true label for sample \( i \) for class \( c \), and \( \hat{y}_{i,c} \) is the predicted probability that sample \( i \) belongs to class \( c \).

\subsubsection{Self-supervised learning} 
Self-supervised learning trains a model without teacher signals. We analyzed two types—contrastive learning and masked image modeling—as these were publicly available supervised fine-tuned versions which enabled comparisons across paradigms (see below).

\paragraph{Contrastive learning}
Contrastive learning encourages similar (positive) pairs to have closer representations while separating dissimilar (negative) pairs. It uses data augmentation and pairwise comparisons to uncover data structures. We examined SimCLR \cite{Chen2020-vo, Chen2020-bn} and SwAV \cite{Caron2020-qh}. SimCLR employs contrastive loss with explicit positive and negative pairs, with the objective:
\[
\mathcal{L}_{\text{SimCLR}}(\theta) = - \sum_{i=1}^{N} \log \frac{ \exp(\text{sim}(f(x_i), f(x_j)) / \tau) }{ \sum_{k=1}^{N} \exp(\text{sim}(f(x_i), f(x_k)) / \tau) }
\]
where \( f(x_i) \) represents the feature vector of sample \( x_i \), \( \text{sim}(\cdot, \cdot) \) is the similarity measure (e.g., cosine similarity), \( \tau \) is a temperature parameter which controls the softness of the distribution, and \( x_j \) is a positive sample which corresponds to \( x_i \) (usually an augmented version of \( x_i \)).
SwAV, used in the SEER model \cite{Goyal2021-kr}, adopts a clustering-based approach, with the objective:
\[
\mathcal{L}_{\text{SwAV}}(\theta) = - \sum_{i=1}^{N} \sum_{c=1}^{C} \text{softmax}(z_i)_{c} \log \hat{q}_{i,c}
\]
where \( z_i \) is the prototype for sample \( i \), \( \hat{q}_{i,c} \) is the predicted cluster assignment for sample \( i \) in cluster \( c \), and \( \text{softmax}(z_i) \) is the softmax output applied to the prototype vector.

\paragraph{Masked Image Modeling (MIM)}
Masked Image Modeling (MIM) trains models to predict masked image regions \cite{He2021-jt, Xie2021-aw}. Randomly masked patches are reconstructed from visible context using an encoder-decoder architecture. For Masked Autoencoders (MAE) \cite{He2021-jt}, the objective is:
\[
\mathcal{L}_{\text{MAE}}(\theta) = \frac{1}{M} \sum_{i \in \mathcal{M}} \| \hat{x}_i - x_i \|^2
\]
where \( x_i \) is the original pixel value of the masked patch, \( \hat{x}_i \) is the predicted pixel value for the masked patch, and \( \mathcal{M} \) denotes the set of indices for the masked patches. \( M \) is the number of masked patches. The loss is the mean squared error between the original and predicted values.

\subsubsection{Supervised fine-tuning after self-supervised pre-training}
We included models trained with self-supervised pre-training followed by supervised fine-tuning, termed self-sup+sup learning. Pre-training uses a self-supervised objective, followed by fine-tuning with a supervised objective (as above).

\subsubsection{Contrastive Language-Image Pretraining (CLIP)}
CLIP employs contrastive learning to align image and text representations in a shared embedding space \cite{Radford2021-ky}. It maximizes similarity between corresponding image-text pairs while minimizing non-corresponding pairs, using weak supervision from natural language descriptions. Its objective is:
\[
\resizebox{\textwidth}{!}{%
    $\mathcal{L}_{\text{CLIP}}(\theta) = - \sum_{i=1}^{N} \log \frac{\exp(\text{sim}(f_\text{I}(x_i), f_\text{T}(t_i)) / \tau)}{\sum_{j=1}^{N} \exp(\text{sim}(f_\text{I}(x_i), f_\text{T}(t_j)) / \tau) + \sum_{j=1}^{N} \exp(\text{sim}(f_\text{I}(x_j), f_\text{T}(t_i)) / \tau)}$
}
\]
where \( f_\text{I}(x_i) \) represents the image encoder’s feature vector for image \( x_i \), \( f_\text{T}(t_i) \) is the text encoder’s feature vector for text \( t_i \), \( \text{sim}(\cdot, \cdot) \) is cosine similarity, and \( \tau \) is a temperature parameter.

\subsubsection{Untrained models}
To provide a baseline, we included untrained, randomly initialized models with architectures ResNet-50, RegNetY-320, and ViT-B/16. To account for initialization randomness, we generated 10 models per architecture and reported the mean, 5th percentile, and 95th percentile of results.

\begin{table}[tp!]
\small
\centering
\resizebox{1.0\textwidth}{!}{
\begin{tabular}{|l|l|l|l|}
\hline
\textbf{Model}             & \textbf{Learning Paradigm} & \textbf{Architecture} & \textbf{Dataset} \\ \hline\hline
Supervised ResNet50 IN1k   & Supervised                & ResNet50              & IN-1k      \\ \hline
Supervised ResNet101 IN1k  & Supervised                & ResNet101             & IN-1k      \\ \hline
Supervised ResNet152 IN1k  & Supervised                & ResNet152             & IN-1k      \\ \hline
Supervised RegNetY320 IN1k & Supervised                & RegNetY320            & IN-1k      \\ \hline
Supervised ViT/B16 IN21k+IN1k & Supervised            & ViT-B/16              & IN-21k + IN-1k \\ \hline
Supervised ViT/B32 IN21k+IN1k & Supervised            & ViT-B/32              & IN-21k + IN-1k \\ \hline
SimCLRv2 ResNet50 IN1k     & Self-Supervised (SimCLR)   & ResNet50              & IN-1k      \\ \hline
SimCLRv2 ResNet101 IN1k    & Self-Supervised (SimCLR)   & ResNet101             & IN-1k      \\ \hline
SimCLRv2 ResNet152 IN1k    & Self-Supervised (SimCLR)   & ResNet152             & IN-1k      \\ \hline
SEER RegNetY320 IG1b       & Self-Supervised (SwAV)     & RegNetY320            & IG-1B     \\ \hline
MAE ViT/B16 IN1k           & Self-Supervised (MIM)      & ViT-B/16              & IN-1k      \\ \hline
SimCLRv2+Sup ResNet50 IN1k  & self-sup+sup (SimCLR)         & ResNet50              & IN-1k      \\ \hline
SimCLRv2+Sup ResNet101 IN1k & self-sup+sup (SimCLR)         & ResNet101             & IN-1k      \\ \hline
SimCLRv2+Sup ResNet152 IN1k & self-sup+sup (SimCLR)         & ResNet152             & IN-1k      \\ \hline
SEER+Sup RegNetY320 IG1b+IN1k & self-sup+sup (SwAV)         & RegNetY320            & IG-1B + IN-1k \\ \hline
MAE+Sup ViT/B16 IN1k        & self-sup+sup (MIM)            & ViT-B/16              & IN-1k      \\ \hline
CLIP ResNet50 CC12M        & CLIP                     & ResNet50              & CC 12M \\ \hline
CLIP ResNet50 OpenAI       & CLIP                     & ResNet50              & OpenAI Dataset   \\ \hline
CLIP ViT/B16 OpenAI        & CLIP                     & ViT-B/16              & OpenAI Dataset   \\ \hline
CLIP ViT/B16 datacomp large & CLIP                   & ViT-B/16              & DataComp L   \\ \hline
CLIP ViT/B16 datacomp xlarge & CLIP                  & ViT-B/16              & DataComp XL  \\ \hline
CLIP ViT/B32 OpenAI        & CLIP                     & ViT-B/32              & OpenAI Dataset   \\ \hline
CLIP ViT/B32 datacomp medium & CLIP                  & ViT-B/32              & DataComp M  \\ \hline
CLIP ViT/B32 datacomp xlarge & CLIP                  & ViT-B/32              & DataComp XL  \\ \hline
Untrained ResNet50          & Untrained                & ResNet50              & -                \\ \hline
Untrained RegNetY320        & Untrained               & RegNetY320           & -                \\ \hline
Untrained ViT/B16        & Untrained               & ViT-B/16            & -                \\ \hline
\end{tabular}
}
\caption{List of models used in the study, including their learning paradigm, architecture, and datasets. The details of each dataset are follows: IN-1k: A subset of ImageNet \cite{Deng2009-zv} introduced in \cite{Russakovsky2015-pn}, IN-21k: The full ImageNet dataset \cite{Deng2009-zv}, IG-1b: The dataset used in \cite{Goyal2021-kr}, CC 12M: Conceptual 12M dataset \cite{Changpinyo2021-yk}, OpenAI Dataset: The dataset used in \cite{Radford2021-ky}, DataComp M, L, XL: Subsets of Datacomp dataset \cite{Gadre2023-na}}
\label{tab:models}
\end{table}

\subsection{Hierarchical clustering analysis}
To evaluate how well the DNN object representations are organized at the coarse category level, we performed hierarchical clustering and compared the results with human-rated coarse categories from the THINGS dataset. We applied hierarchical clustering to embeddings derived from each model’s latent representations using the \texttt{scipy.cluster.hierarchy.linkage} function with the Ward linkage method.

To quantitatively compare the clustering results of DNNs to the human coarse categories, we calculated the adjusted mutual information (AMI) \cite{Vinh2009-fd}, defined as:

\[
\text{AMI}(U, V) = \frac{I(U, V) - \mathbb{E}[I(U, V)]}{\max(H(U), H(V)) - \mathbb{E}[I(U, V)]}
\]

where \(I(U, V)\) is the mutual information between the clustering results of DNNs and human coarse categories, \(H(U)\) and \(H(V)\) are the entropies of the DNN clustering and human-rated coarse categories, respectively, and \(\mathbb{E}[I(U, V)]\) is the expected mutual information, correcting for chance agreement. Higher AMI scores indicate stronger alignment between DNN clustering results and human-rated coarse categories.

\section{Results}
\subsection{Supervised comparison shows limited differentiation of human-DNN similarity across learning paradigms}
We first examined the similarity between human and Deep Neural Network (DNN) object representations across different learning paradigms using Representational Similarity Analysis (RSA), a conventional supervised comparison method that precedes our primary focus on unsupervised alignment. Using the THINGS dataset, we computed Pearson correlations between human representational dissimilarity matrices (RDMs) and those from 27 DNN models, including supervised learning, self-supervised learning, self-sup+sup learning, and CLIP models (see Methods).

\begin{figure}[tp!]
\centering
\includegraphics[width=\textwidth]{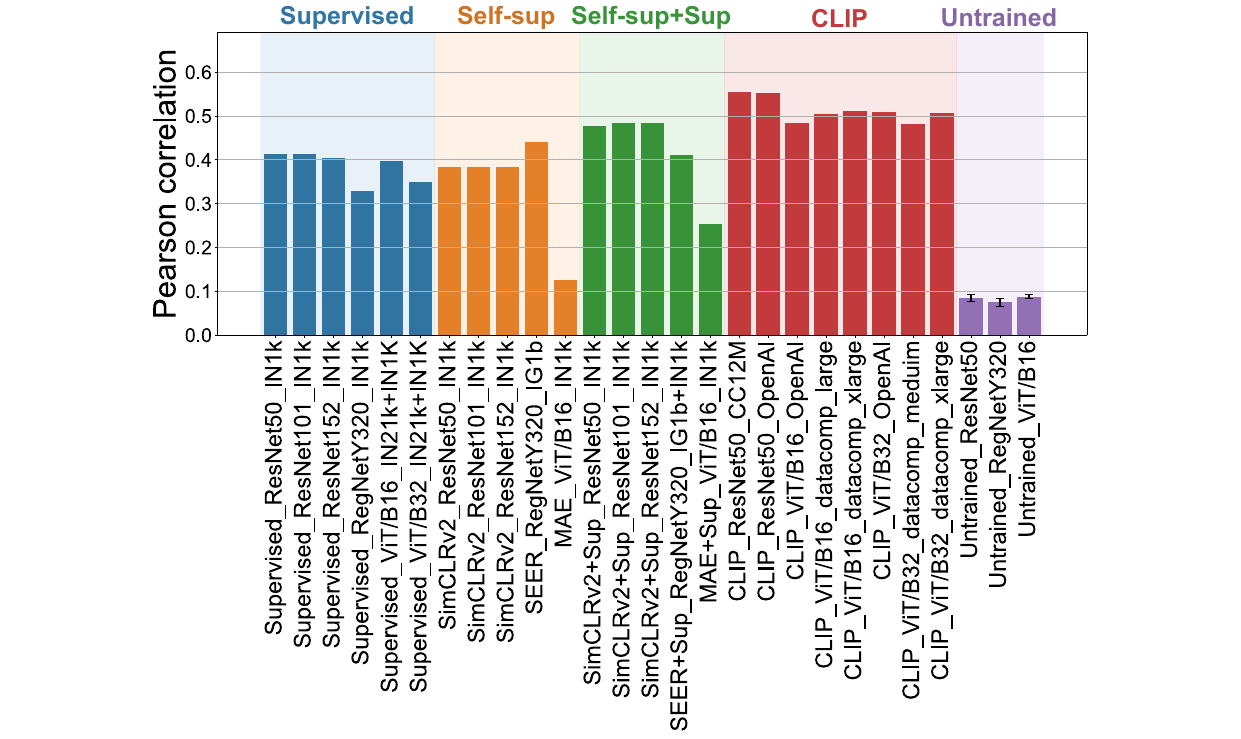}
\caption{\textbf{Supervised comparison between representational structures of Humans and DNNs. }Pearson correlation coefficients between human and model RDMs
based on conventional Representational Similarity Analysis (RSA). The results for the untrained models are averaged over 10 random initializations, with error bars representing the 5th and 95th percentiles of these trials. }
\label{fig:RSA}
\end{figure}

RSA results showed that most DNN models achieved moderate to high correlations with human representations, yet revealed little distinction across different learning paradigms (Figure \ref{fig:RSA}). Specifically, both supervised learning models and self-supervised learning models using contrastive learning (i.e., SimCLR and SEER models) yielded correlations of around 0.4 with negligible differences between them, while all outperformed untrained models. Self-sup+sup models exhibited slightly higher correlations (around 0.45), and CLIP-based models performed marginally better (around 0.5), though these differences were small and not clearly distinguishable. Some exceptions were observed in the MAE (self-supervised) and the MAE+Sup (self-sup+sup) models, which produced much lower correlations, about 0.1 and 0.25 respectively. Taken together, these findings suggest that while RSA captures broad similarities, its reliance on fixed mappings limits its ability to distinguish between coarse-grained and fine-grained structural correspondence.

\subsection{Unsupervised Alignment and learning paradigm differences in fine-grained human-DNN representation matching}
To evaluate how models trained with different learning paradigms align with human object representations at the fine-grained level, we conducted unsupervised alignment using Gromov-Wasserstein Optimal Transport (GWOT) on the THINGS dataset (see Method). As in the supervised comparison, this analysis compared human representations with those of 27 DNN models across four learning paradigms: supervised, self-supervised, self-sup+sup, and CLIP. From the unsupervised alignment, we derived optimal transportation plans between human and model RDMs and evaluated alignment by calculating matching rates.

\subsubsection{Representative examples of unsupervised alignment results between humans and models}
To illustrate our analyses before presenting comprehensive results for all models, we highlight one model that aligns well with human representations through unsupervised alignment and one that does not. Specifically, we show CLIP ViT/B16 datacomp large as an example of a well-aligned model and SimCLRv2 ResNet50 IN1k as an example of a poorly aligned model.

\begin{figure}[tp!]
\centering
\includegraphics[width=\textwidth]{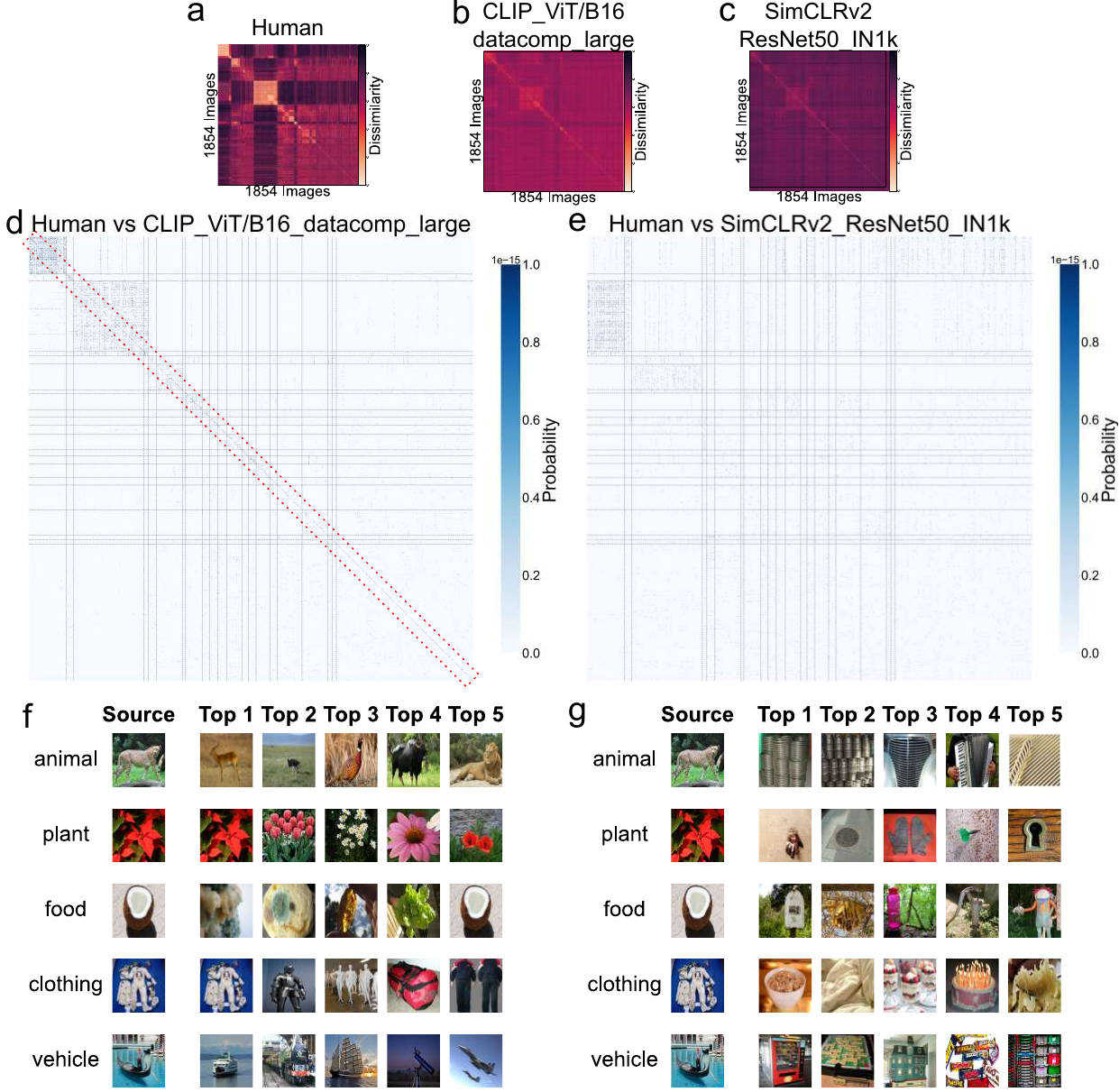}
\caption{\textbf{Representative Example of Unsupervised Alignment Results Between Human and Model.}
 Representational Dissimilarity Matrix (RDM) of \textbf{a.}human, 
\textbf{b.} CLIP ViT/B16 datacomp large, and
\textbf{c.} SimCLRv2 ResNet50 IN1k. Objects in RDMs are sorted by coarse categories.
\textbf{d.} Optimal transportation plan between human and CLIP ViT/B16 datacomp large. Rows represent human objects, and columns represent model objects. Objects are sorted by coarse categories. The values indicate transportation probabilities, with prominent diagonal patterns reflecting strong alignment (surrounded by a red dashed line).
\textbf{e.} Optimal transportation plan between human and SimCLRv2 ResNet50 IN 1k. Unlike d, the values of the diagonal components are small. 
\textbf{f.} Examples of top 5 objects with the highest transportation probability between human and CLIP ViT/B16 datacomp large.
\textbf{g.} Examples of top 5 objects with the highest transportation probability between human and SimCLRv2 ResNet50 IN 1k.
}
\label{fig:representative_gwot}
\end{figure}

As an example of successful alignment, CLIP ViT/B16 datacomp large achieves a matching rate of 26.2\% out of 1,854 objects with humans. We performed unsupervised alignment using GWOT between the RDMs shown in Figures \ref{fig:representative_gwot}a and b. The success of this alignment is reflected in the optimal transportation plan having a roughly diagonal pattern (Figure \ref{fig:representative_gwot}d) and the transported objects (Figure \ref{fig:representative_gwot}f). The optimal transportation plan in Figure \ref{fig:representative_gwot}d represents the optimal solution with the lowest Gromov-Wasserstein Distance (GWD) among the solutions found in the iterative optimization process. The roughly diagonal pattern of this matrix indicates that the same objects are predominantly transported between humans and CLIP. We can also observe this success in matching in Figure \ref{fig:representative_gwot}f, which highlights the top-5 objects with the highest transportation probabilities. We found that the same object as the source object is often in the Top 5, and even when incorrect, it is likely to be transported to semantically similar objects. These findings demonstrate that CLIP effectively captures the fine-grained structure of human object representations.

In contrast, SimCLRv2 ResNet50 IN1k fails to align, showing a matching rate of 0\% for 1,854 objects with human representations. We performed unsupervised alignment using GWOT between the RDMs shown in Figures \ref{fig:representative_gwot}a and c. Unlike CLIP, the optimal transportation plan lacks a clear diagonal pattern (Figure \ref{fig:representative_gwot}e), indicating inconsistent object correspondences between humans and SimCLR representations. This misalignment is further evident in Figure \ref{fig:representative_gwot}g, which highlights the top-5 transported objects, showing that the source object is rarely matched to itself or semantically similar objects. These findings demonstrate that the SimCLR model's object representations lack the fine-grained structure of human object representations.

\subsubsection{Unsupervised alignment results between humans and all models}
\begin{figure}[tp!]
\centering
\includegraphics[width=\textwidth]{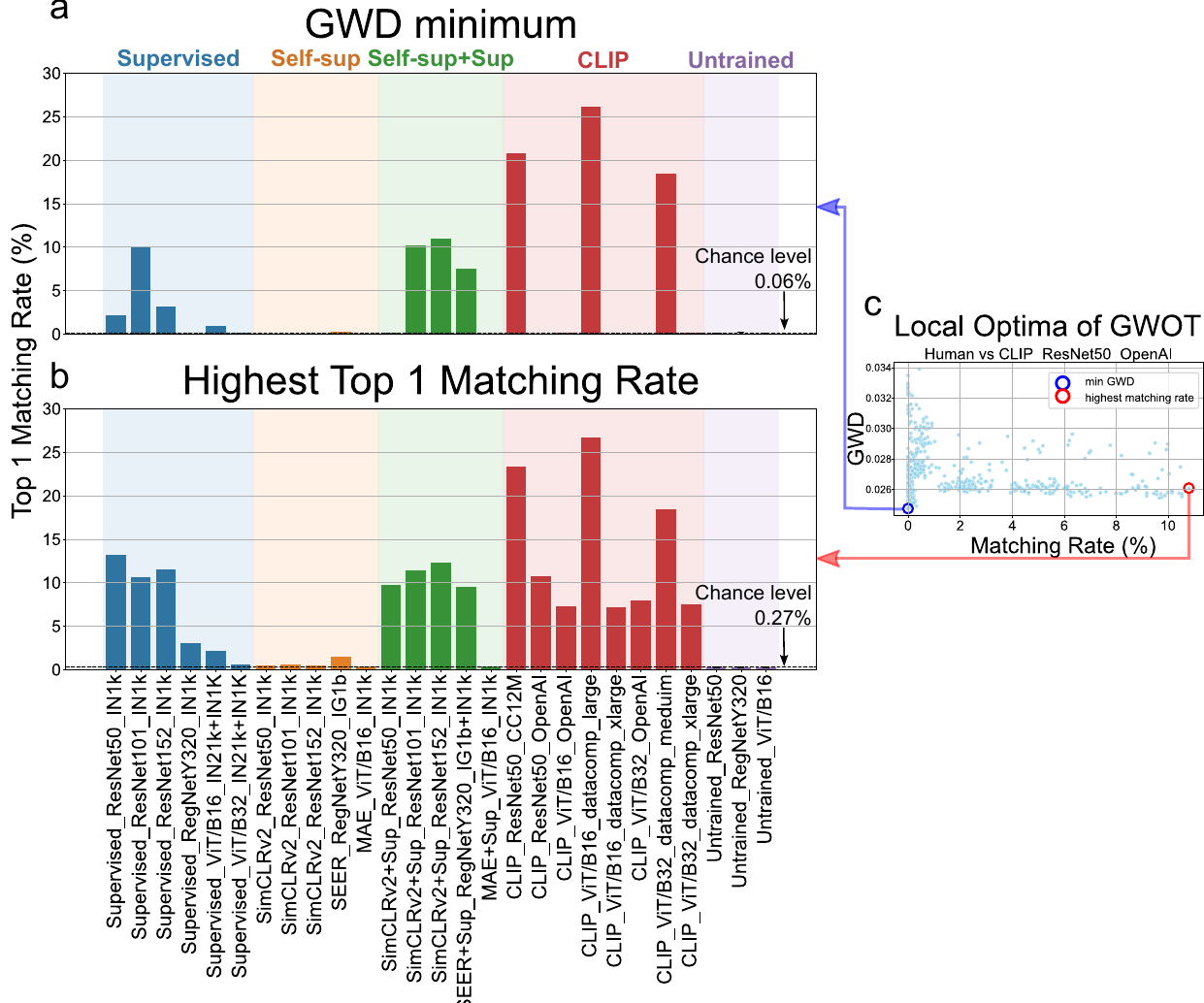}
\caption{\textbf{Fine-grained matching of representational structures between humans and DNNs.}
\textbf{a.} Matching rates between human and model object representations for the minimum GWD solution based on GWOT. The results for the untrained model are averaged over 10 random initializations, with error bars representing the 5th and 95th percentiles of these trials. Simulation results for the chance level are also shown as error bars (5th and 95th percentiles), but both error bars are too small to be visible in this plot. This solution is indicated by the blue circle in panel c.
\textbf{b.} Matching rates between human and model object representations for the highest matching rate solution based on GWOT. Similar to panel a, the untrained model results are averaged over 10 random initializations, with error bars representing the 5th and 95th percentiles. Chance level simulation results are also shown as error bars (5th and 95th percentiles), but both error bars are too small to be visible in this plot. This solution is indicated by the red circle in c.  
\textbf{c.} A representative example of local optima from GWOT optimization (human vs. CLIP ResNet50 OpenAI). Each point represents a transportation plan obtained from the optimization results.
}
\label{fig:object_level}
\end{figure}

We evaluate the unsupervised alignment between human representations and 27 DNNs by calculating the matching rate for both the solution with the minimum GWD (Figure \ref{fig:object_level}a) and the local optimal solution that achieves the highest matching rate (Figure \ref{fig:object_level}b). The minimum-GWD solution represents the lowest point in the iterative optimization process (blue circle in Figure \ref{fig:object_level}c). as this solution may be influenced by chance due to finite optimization steps, we also evaluated the solution with the highest matching rate, which reflects the best matching rate achieved during the process (red circle in Figure \ref{fig:object_level}c; see Method).

The matching rates for the minimum GWD transportation plans (Figure \ref{fig:object_level}a) reveal a clear distinction, compared with the results based on the conventional RSA (Figure \ref{fig:RSA}): some models trained with linguistic information (i.e., supervised, self-sup+sup, and CLIP) achieve significantly higher matching rates, while no self-supervised models achieve a matching rate above the chance level. Notably, certain CLIP models, such as CLIP ResNet50 CC12M, CLIP ViT/B16 datacomp large, and CLIP ViT/B32 datacomp medium, performed especially well, reaching matching rates of around 20\% of 1,854 objects. Meanwhile, some supervised models (e.g., Supervised ResNet101 IN1k) achieved matching rates of around 10\% of 1,854 objects, as did self-sup+sup models (e.g., SimCLRv2+Sup ResNet101 IN1k, SimCLRv2+Sup ResNet152 IN1k), which, after supervised fine-tuning, significantly outperformed self-supervised-only models at the chance level.

Differences between models are further clarified by evaluating the highest matching rates among local optima (Figure \ref{fig:object_level}b) as well as minimum GWD solutions (Figure \ref{fig:object_level}a). At minimum GWD (Figure \ref{fig:object_level}a), except for the well-aligned models, even those using linguistic information achieve mostly chance-level matching rates, showing little difference from self-supervised models also at the chance level. In contrast, the transportation plans with the highest matching rates reveal that most of the linguistic models including supervised, self-sup+sup and CLIP models, exhibit consistently high matching rates around 10\%, well above the chance level, while self-supervised models remain at the chance level, with even the best performing model, SEER RegNetY320 IG1b, reaching only 1.46\%. Taken together, these results based on the highest matching rates and minimum GWD solutions (Figures \ref{fig:object_level}a,b) suggest that linguistic information enables fine-grained matching with human object representations in models that cannot be achieved by self-supervised learning alone.

\subsection{Unsupervised alignment and learning paradigm differences in coarse-grained human -DNN representation matching}
\begin{figure}[tp!]
\centering
\includegraphics[width=\textwidth]{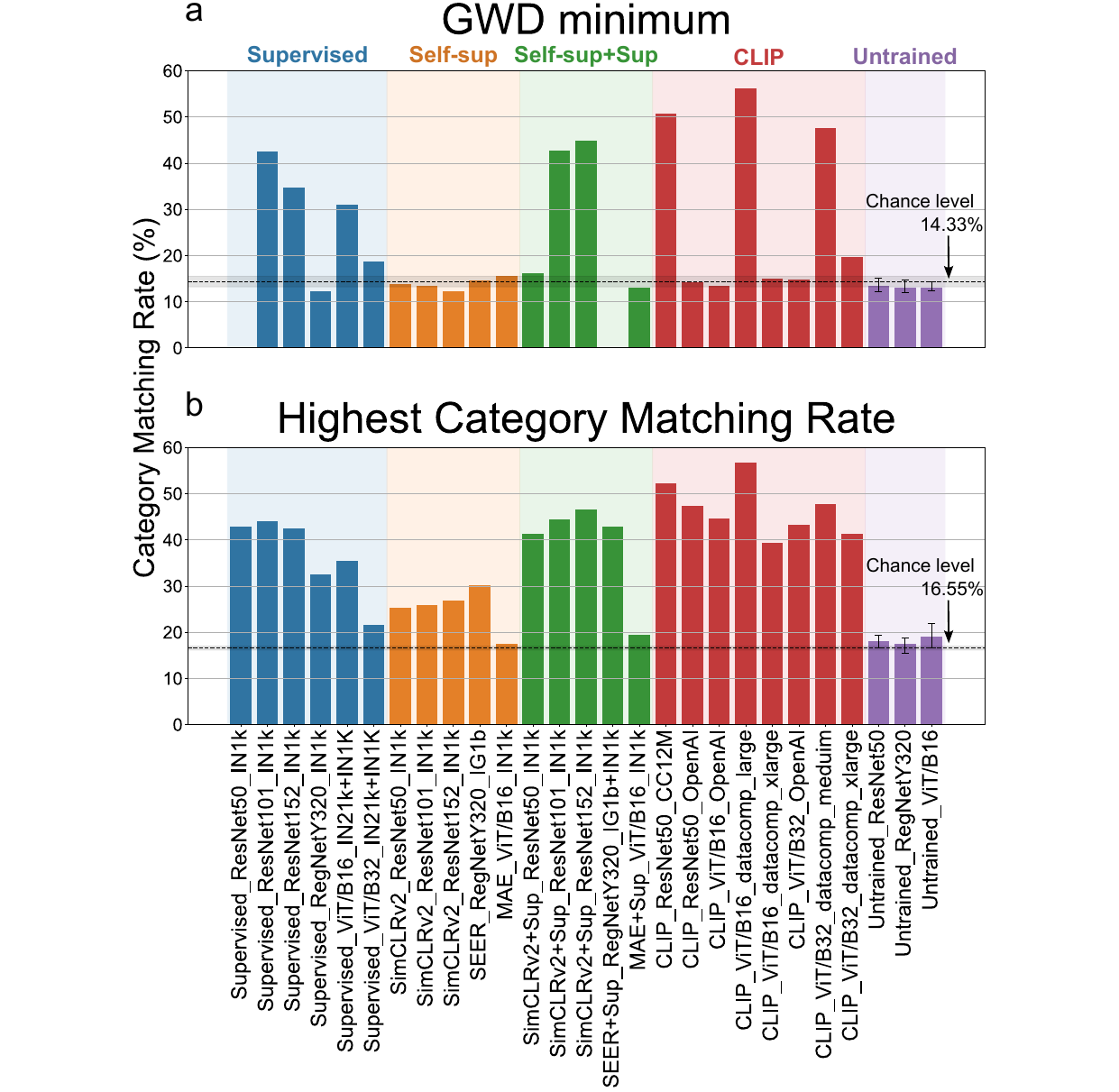}
\caption{\textbf{Coarse-grained matching of representational structures between humans and DNNs.}
\textbf{a.} Category Matching rates between human and model object representations for the minimum GWD solution based on GWOT. Results for the untrained model are averaged over 10 random initializations, with error bars representing the 5th and 95th percentiles of these trials. Simulation results for the chance level are also shown as error bars (5th and 95th percentiles).
\textbf{b.} Category Matching rates between human and model object representations for the highest category matching rate solution based on GWOT. Similar to panel a, the untrained model results are averaged over 10 random initializations, with error bars representing the 5th and 95th percentiles, and the chance level simulation results are also shown as error bars (5th and 95th percentiles)}
\label{fig:coarse-grained}
\end{figure}

To examine how models trained with different learning paradigms align with human object representational structures at the coarse-grained level, we evaluated the results of unsupervised alignment using the category matching rate. This metric considers a mapping as “correct” if the mapped objects belong to the same category. For category definitions, we used the coarse categories defined by human raters in the THINGS dataset (see Methods). Similar to the evaluation of fine-grained matching, we report the category matching rate for both the minimum GWD solution (Figure \ref{fig:coarse-grained}a) and the local optimal solution with the highest category matching rate (Figure \ref{fig:coarse-grained}b). 

The category matching rates for minimum GWD transportation plans (Figure \ref{fig:coarse-grained}a) revealed that models trained with linguistic information, which exhibited fine-grained matching with humans, also achieved strong coarse-grained matching, whereas self-supervised models struggled to align even at this level. As in the fine-grained results, certain CLIP models (e.g., CLIP ResNet50 CC12M, CLIP ViT/B16 datacomp large, and CLIP ViT/B32 datacomp medium) performed particularly well, achieving category matching rates of around 50\%. Some supervised and self-sup+sup models (e.g., Supervised ResNet101 IN1k, SimCLRv2+Sup ResNet101 1k, and SimCLRv2+Sup ResNet152 IN1k) also exceeded untrained models and chance levels, reaching category matching rates above 40\%. In contrast, all self-supervised models that did not utilize linguistic information remained at the chance level (around 15\%),while self-sup+sup models benefited significantly from supervised fine-tuning, achieving higher rates.

Evaluating the highest category matching rates among local optima, together with the minimum GWD solution, highlights the robust coarse-grained matching achieved by models trained with linguistic information. At minimum GWD (Figure \ref{fig:coarse-grained}a), except for well-aligned models like CLIP ResNet50 CC12M achieving around 50\%, other linguistic models remain at a chance level, showing little difference from chance level or untrained models. In contrast, transportation plans with the highest category matching rates reveal that most of these linguistic models exceed 40\%, far surpassing untrained baselines and chance level.

Furthermore, even self-supervised models using contrastive learning show sightly above-chance level category matching rates (Figure \ref{fig:coarse-grained}b) when evaluated at the highest matching rates, hinting at potential coarse-grained structure in their representations. At minimum GWD solutions (Figure \ref{fig:coarse-grained}a), all self-supervised models remain at chance levels, showing no coarse-grained matching with human object representations. However, at the highest category matching rates, models trained with self-supervised contrastive learning (i.e., SimCLR and SEER models) reach category matching rates of around 25\%, exceeding chance and untrained baseline. In contrast, the MAE model, also a self-supervised model, remains at the chance level (around 15\%), highlighting variations within self-supervised learning approaches. This improvement in contrastive learning models, despite the limitations shown by models like MAE, suggests that certain self-supervised learning approaches may capture basic category-level distinctions, a hypothesis we explore further through clustering analysis in the next section.

\subsection{Clustering and coarse human-like categories across learning paradigms in DNN representations}

\begin{figure}[tp!]
\centering
\includegraphics[width=\textwidth]{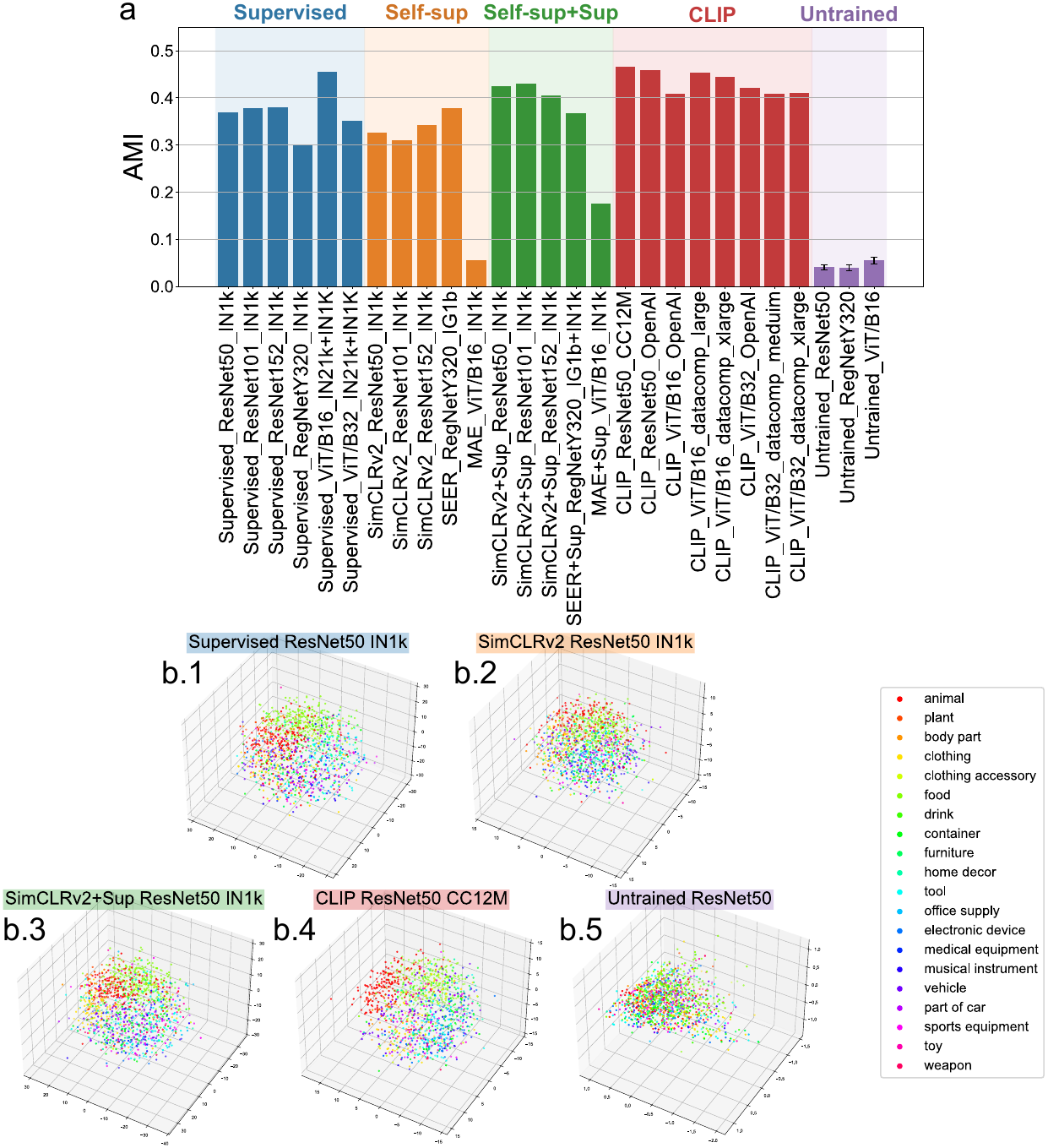}
\caption{\textbf{Hierarchical clustering and MDS visualization of model object representations}
\textbf{a.} Adjusted Mutual Information (AMI) scores between the clusters of model object representations and the coarse categories in the THINGS dataset, showing the alignment of model clusters with human-labeled categories. The results for the untrained model are averaged over 10 random initializations, with error bars representing the 5th and 95th percentiles of these trials. 
\textbf{b.} Multi-Dimensional Scaling (MDS) visualization of model object representations colored by their corresponding THINGS coarse categories.}
\label{fig:clustering}
\end{figure}

To directly test whether the above chance-level performance of self-supervised models using contrastive learning at the highest category matching rates (Figure \ref{fig:coarse-grained}b) and the moderate correlations observed in RSA (Figure \ref{fig:RSA}) reflect the formation of coarse human-like category structures, we applied hierarchical clustering to model embeddings and compared the resulting clusters with human-labeled coarse categories from the THINGS dataset. While the category matching rates obtained via GWOT suggest possible alignment at the coarse category level, and the RSA results suggest representational similarity driven by coarse-grained structures (as noted in the Introduction), neither of these results directly demonstrates that the models organize objects into human-like coarse categories. This clustering analysis addresses this gap by explicitly evaluating category formation, with correspondence quantified using Adjusted Mutual Information (AMI).

As shown in Figure \ref{fig:clustering}a, self-supervised models using contrastive learning achieved AMI scores significantly higher than those of untrained models, comparable to supervised learning models. Self-sup+sup models based on contrastive learning slightly outperformed their self-supervised counterparts, suggesting that supervised fine-tuning enhances category formation, albeit that self-supervised models alone achieved scores nearly as high. Consistent with the highest category matching rates (Figure \ref{fig:coarse-grained}b), MAE and MAE+Sup models yielded low AMI scores, highlighting variations within self-supervised learning paradigms.

We also visualized the embeddings of these models using Multidimensional Scaling (MDS), and found that self-supervised models formed distinct clusters when colored by THINGS coarse categories (Figure \ref{fig:clustering}b.2), unlike the unstructured embeddings of untrained models (Figure \ref{fig:clustering}b.5). Models trained with linguistic information (e.g., Figure \ref{fig:clustering}b.1, b.3, b.4) similarly showed clear clustering, reflecting strong alignment with human categories. Taken together, these findings from clustering and visualization demonstrate that self-supervised learning, particularly with contrastive methods, can effectively capture coarse human-like category structures even without linguistic information.

\section{Discussion}
We demonstrated that CLIP models, which leverage structured linguistic information, achieved the strongest fine-grained alignment with human object representational structures, and significantly outperformed other learning paradigms (Figure \ref{fig:object_level}). Supervised and self-sup+sup models also showed above-chance fine-grained matching rates. In contrast, self-supervised models failed to align at this level, suggesting that fine-grained human-like object representations critically depend on linguistic information, particularly the structured semantic knowledge utilized by CLIP.

Although self-supervised learning did not align with human representations at fine-grained level, it modestly captured coarse human-like category structures, especially through contrastive methods, as shown by category matching and clustering analyses. At the highest category matching rates (Figure \ref{fig:coarse-grained}b), self-supervised models using contrastive learning achieved slightly above-chance rates. Clustering results (Figure \ref{fig:clustering}a) further confirmed that these models formed clusters which broadly corresponded with human coarse categories, indicating that self-supervised learning can independently capture coarse categorical structures without linguistic input.

\subsection{Self-supervised learning as a model of Prelinguistic Category Formation}
Self-supervised learning may shape the broad categorical structure of object representations in prelinguistic infants. Our findings show that self-supervised models, particularly those using contrastive learning, form coarse category clusters that modestly align with human categories. This suggests that self-supervised learning, by detecting visual patterns without external supervision, mirrors how infants organize their environment into coarse categories using perceptual cues before acquiring language. Previous studies have shown that infants can learn statistical regularities in visual input early in development \cite{Kirkham2002-nv, Fiser2002-pn, Bulf2011-yg}, supporting the idea that self-supervised mechanisms play a key role in early cognition. Recent computational work also indicates that self-supervised models trained on natural images can recover high-level category structure that resembles human conceptual groupings \cite{Zhuang2021-ek, Prince2024-zf}. These prior studies and our findings support the hypothesis that infants use self-supervised learning to build a foundational model for later cognition during their prelinguistic period \cite{Cusack2024-yw}. Future research could directly test this by comparing model behavior with infant developmental data.

\subsection{Implications of semantic similarity in the THINGS dataset}
The THINGS dataset’s odd-one-out task is designed to require participants to identify the odd object among three distinct object images, and thereby encourage decisions based on semantic similarities rather than visual appearance \cite{Mahner2024-kh}. We argue that this focus on semantics favors models like CLIP and supervised models, which leverage visual and linguistic modalities for nuanced comprehension. In contrast, we argue that self-supervised models, which rely solely on visual patterns, perform near chance, reflecting their challenges in capturing conceptual relationships without linguistic input.

\subsection{Limitations}
While our study investigated the impact of different learning paradigms on alignment with human object representational structures, the characteristics of the dataset used for training and model architecture can also be significant factors in shaping such alignment \cite{Mehrer2021-cj, muttenthaler2023-zf, Conwell2024-mp}. We acknowledge that the high computational cost of Gromov-Wasserstein Optimal Transport limited the number of models we could evaluate, constraining our analysis to learning paradigms alone. Consequently, we were unable to systematically control for the effects of architectural variations or dataset-specific factors, such as image diversity or semantic granularity.

\subsection*{CRediT authorship contribution statement}
\textbf{Soh Takahashi:} Conceptualization, Methodology, Formal analysis, Writing - Original Draft, Writing - Review \& Editing. \textbf{Masaru Sasaki:} Conceptualization, Methodology, Formal analysis, Writing - Review \& Editing. \textbf{Ken Takeda:} Methodology, Writing - Review \& Editing. \textbf{Masafumi Oizumi:} Conceptualization, Methodology, Writing - Original Draft, Writing - Review \& Editing.

\subsection*{Declaration of competing interest}
The authors declare no competing interests

\subsection*{Acknowledgments}
MO was supported by JST Moonshot R\&D Grant Number JPMJMS2012 and Japan Promotion Science, Grant-in-Aid for Transformative Research Areas Grant Number 23H04834.

\subsection*{Declaration of Generative AI and AI-assisted technologies in the writing process}
During the preparation of this work, the authors used ChatGPT (OpenAI) in order to assist with improving the clarity and fluency of the English text. After using this tool, the author reviewed and edited the content as needed and take full responsibility for the content of the publication.

\bibliography{main}
\bibliographystyle{unsrt}
\end{document}